\documentclass[graybox]{svmult}
\usepackage{type1cm}        
\usepackage{makeidx}         
\usepackage{graphicx}        
\usepackage{multicol}        
\usepackage[bottom]{footmisc}
\usepackage{multirow}
\usepackage{amsmath}
\usepackage{newtxtext}       %
\usepackage{newtxmath}       
\usepackage{booktabs}

\makeindex            

\begin{document}

\title*{Federated and Transfer Learning: A Survey on Adversaries and Defense Mechanisms}

\titlerunning{Federated and Transfer Learning: A Survey on Adversaries and Defense Mechanisms} 
\author{Ehsan Hallaji, Roozbeh Razavi-Far and Mehrdad Saif}
\institute{Ehsan Hallaji \at Department of Electrical and Computer Engineering, University of Windsor, 401 Sunset Avenue, Windsor, ON N9B 3P4, Canada, \email{hallaji@uwindsor.ca}
\and Roozbeh Razavi-Far \at Department of Electrical and Computer Engineering and School of Computer Science, University of Windsor, 401 Sunset Avenue, Windsor, ON N9B 3P4, Canada, \email{roozbeh@uwindsor.ca}
\and Mehrdad Saif \at Department of Electrical and Computer Engineering, University of Windsor, 401 Sunset Avenue, Windsor, ON N9B 3P4, Canada, \email{msaif@uwindsor.ca}}

\maketitle

\abstract{The advent of federated learning has facilitated large-scale data exchange amongst machine learning models while maintaining privacy. Despite its brief history, federated learning is rapidly evolving to make wider use more practical. One of the most significant advancements in this domain is the incorporation of transfer learning into federated learning, which overcomes fundamental constraints of primary federated learning, particularly in terms of security. This chapter performs a comprehensive survey on the intersection of federated and transfer learning from a security point of view. The main goal of this study is to uncover potential vulnerabilities and defense mechanisms that might compromise the privacy and performance of systems that use federated and transfer learning.}

\section{Introduction}
Machine learning has exploded in popularity as the information era has matured. As a sub-discipline of machine learning, deep learning (DL) is responsible for a number of achievements that have helped popularise the area. The hierarchical feature extraction within the DL models enables them to learn complex underlying patterns in the observed input space. This makes DL models suitable for processing various data types and facilitating different tasks 
such as prediction \cite{HASSANI2022118861}, detection \cite{8608001}, imputation \cite{9370000}, and data reduction \cite{HASSANI2021104150}.
Although the success of DL-based projects is contingent on several factors, one of the most important requirements is usually to have access to abundant training samples.

With the advancement of technologies such as internet of things and increasing number of intelligent devices, the diversity and volume of generated data is growing at an astonishing pace \cite{9608954}. This abundant data stream is dispersed and diverse in character. When this data is evaluated as a whole, it may provide knowledge and discoveries that might possibly accelerate technological and scientific advancements. Nonetheless, the privacy hazards associated with data ownership are increasingly becoming a major problem. The conflict between user privacy and high-quality services is driving need for new technologies and research to enable knowledge extraction from data without jeopardising data-holding parties' privacy.

Federated learning (FL) is perhaps the most recent approach presented to potentially resolve this issue. FL allows for the collaborative training of a DL model across a network of client devices. This multi-party collaboration is accomplished by communication with a central server and decentralisation of training data \cite{pmlr-v54-mcmahan17a}. In other words, the FL design requires that the training data be retained on the client, or local device, that generated or recorded it. FL's data decentralisation addresses a significant portion of data and user privacy concerns, since model training at the network's edge eliminates the need for direct data sharing. Nevertheless, in FL systems one should strike a balance between data privacy and model performance. The appropriate balance is determined by a number of criteria, including the model architecture, data type, and intended application. Beyond FL restrictions, ensuring the confidentiality and security of the FL infrastructure is critical for establishing trust amongst diverse clients of the federated network.

The conventional FL imposes a constraint by requiring customers' training data to use similar attributes. However, most industries such as banking and healthcare do not comply with this assumption. In centralized machine learning, this was addressed by Transfer Learning (TL), which enables a model obtained from a specific domain to be used in other domains of with the same application \cite{8608001}. Inspired by this, Federated TL (FTL) emerged as a way to overcome this constraint \cite{9076003}. FTL clients may differ in the employed attributes, which is more practical for industrial application. TL effectiveness is heavily reliant on inter-domain interactions. It is worthwhile to mention that organizations joined in a FTL network are often comparable in the service they provide and the data they use. As a result, including TL within the FL architecture can be quite advantageous.

FTL is at the crossroads of two distinct and rapidly expanding research areas of privacy-preserving and distributed machine learning. For this reason, it is crucial to investigate more on these two topics to make best use of FTL. Hence, this chapter studies different security and privacy aspects of FTL. 

Information privacy and machine learning are distinct and rapidly expanding research areas that derive FTL, and, thus, going through their connections to FTL is necessary. Understanding the interplay between TL and FL, as well as identifying potential risks to FTL in real-world applications, is crucial. Knowing compatible diffense mechanisms with FTL is also vital for mitigating potential cyber-threats. Hence, in this chapter, we present a comprehensive survey on possible threats to FTL and the available defense mechanisms. 

The rest of the chapter is organized as follows. The preliminaries of this survey are explained in Section 2. Section 3 reviews known attack scenarios on FL and TL w.r.t. performance and privacy. Section 4 presents tools and defense mechanisms that are undertook in the literature for mitigating threats to FL and TL. Section 5 explains the future directions in defending FTL. Finally, Chapter 6 concludes the conducted survey.

\section{Background}
This chapter concisely reviews preliminaries of federated and transfer learning to facilitate the discussions in the following sections.

\subsection{Federated Learning}
FL paradigm allows for collaborative model training across several participants (i.e., also referred to as clients or devices). This multi-party collaboration is accomplished by communication with a central server and decentralization of training data \cite{pmlr-v54-mcmahan17a}. Data decentralization of FL mitigates a major part of user privacy issues. Moreover, the efficiency of FL reduces the communication overhead in the network. FL can be categorizes from different perspectives, as explained in the following.

\subsubsection{Categories of Federated Learning}
FL variations generally fall under three categories depending on the portion of feature and sample space they share \cite{10.1145/3298981}:
\begin{enumerate}
\item \textbf{Horizontal Federated Learning:} Participants exchange data with comparable properties captured from various users \cite{1316832}. For instance, clinical information of various patients is recorded using the same features across several hospitals. Therefore, similar deep learning structures can be trained on these datasets since they all process the same number and types of features.

\item \textbf{Vertical Federated Learning:} The vertical variation is utilized in applications where participant datasets have considerable overlaps in the sample space but each has a separate set of attributes \cite{10.1145/775047.775142}.

\item \textbf{Federated Transfer Learning:} FTL facilitates knowledge transfer between participants when the overlap between sample and feature space is minimal \cite{9076003}. FTL is discussed in further detail later in this section.
\end{enumerate}

\subsection{Transfer Learning}
Most machine learning approaches work on the premise that the training and test data are in the same feature space. In industry, data may be hard to collect in certain applications, and, thus, there is a preference for using the available data shared by large companies and organizations. The challenge, however, is the difference between the data distributions despite the similarity of the applications. In other words, the ideal model needs to model from one domain with limited data resources, while abundant training data is available in another domain. For instance, consider a factory that trained a model to predict the market demand to adjust it production rate for different items. However, it may be time consuming to obtain enough samples for each produced item separately. Instead, TL can enable the model to be trained on the data of other organizations that produce similar products, albeit samples may not be recorded using the same features. The past decade has witnessed an increasing attraction towards research on TL, which resulted in proposing different variations TL under different names \cite{5288526}. 

\subsection{Federated Transfer Learning}
Similar to TL, clients of FTL may not use the same attributes in the training data. This is mostly the case in organizations that are similar in nature but are not identical \cite{DBLP:journals/corr/abs-2010-15561}. Due to such differences, these organizations share only a small portion of each other's feature space. Therefore, under this condition, both samples and features are different in the dataset. Note that the considered condition in FTL is in contrast to the other variants of FL.

FTL takes a model that has been constructed on source data, and then aligns it to be employed in a target domain. This allows the model to be utilized for uncorrelated data points while exploiting the information gained from non-overlapping features in the source domain. As a result, FTL transmits information from the source domain's non-overlapping attributes to new samples within the target domain.

Existing literature on FTL mainly studies the customization of FTL for certain applications \cite{9076082,9099064}. From a learning stand-point, only a limited number of works present distinct FTL protocols. A secure FTL framework is proposed in \cite{9076003} that uses secret sharing and homomorphic encrypton to protect privacy without sacrificing accuracy, which is a typical issue in privacy-preserving techniques. Other benefits of this method include the simplicity of homomorphic encryption and the fact that secret sharing ensures zero accuracy loss and quick processing time. On the other had, Homomorphic encryption, imposes significant computing overhead, and secret sharing necessitated offline operations prior to online computation. A following research \cite{9006280} tackles the computation overhead of previous protocols and extends the FTL model beyond the semi-honset setting by taking malicious users into count as well. The authors use secret-sharing in the designed algorithm to enhance the security and efficiency of multi-party communications in FTL. An scalable heterogeneous FTL framework is also presented in \cite{9005992}, which uses secret sharing and homomorphic encryption.

\section{Threats to Federated Learning}

\begin{table*}[b]
\centering
\setlength{\tabcolsep}{15pt}
\caption{Identification of sources of attacks on FL systems.}
\begin{tabular}{ll}
\toprule
Attacks & Source of Attack \\
\midrule
Data poisoning & Malicious client \\
Model poisoning & Malicious client\\
Backdoor attack & Malicious client and malicious server\\
Evasion attack & Malicious client and model deployment\\
Non-robust aggregation & Aggregation algorithm\\
Training rule manipulation & Malicious client\\
Inference attacks & Malicious server and communication\\
GAN reconstruction & Malicious server and communication\\
Free-riding attack & Malicious client\\
Man-in-the-middle attack & Communication\\
\bottomrule
\end{tabular}
\label{tab:tab1}
\end{table*}

Threats to FL and TL often compromise the functionality or privacy of the system. Table \ref{tab:tab1} lists the sources of threats for common attacks to FL. FL enables a distributed learning process without requiring data exchange, allowing members to freely join and exit federations. Recent research has shown, however, that resilience of FL against the mentioned threats in Table \ref{tab:tab1} can be questionable.

Existing FL protocol designs are vulnerable to rough servers and adversarial parties. While both infer confidential information from participants' updates, the former mainly tampers with the model training whereas the latter resorts to poisoning attacks to deviate the aggregation procedure.

During the training process, communicating model updates might divulge critical information and lead to deep leakage. This can consequently jeopardize the privacy of local data or lead to high-jacking the training data \cite{NEURIPS2019_60a6c400}. The robustness of FL systems, on the other hand, can be degraded using poisoning attacks on the model to corrupt the model or training data \cite{DBLP:journals/corr/abs-2007-05084,pmlr-v108-bagdasaryan20a, pmlr-v97-bhagoji19a}. In turn, these attacks lead to planting a backdoor into the global model or degrade its convergence.

\subsection{Threat Models}
Attacks on FL can be launched in different fashions \cite{DBLP:journals/corr/abs-2012-06337, https://doi.org/10.48550/arxiv.2003.02133}. To have a better grasp of the nature of FL attacks, we will first go through the most frequent threat models in the following:

\begin{itemize}
\item \textbf{Outsider Adversaries} include attacks by eavesdroppers on the line of communication between clients and the FL server, as well as attacks by clients of the FL model once it is provided as a service.
\item \textbf{Insider Adversaries} involve attacks initiated from the server or the edge of the network. Byzantine \cite{NIPS2017_f4b9ec30} and Sybil \cite{DBLP:journals/corr/abs-1808-04866} attacks can be mentioned as two most important insider attacks.  
\item \textbf{Semi-Honest Adversaries} are non-aggressive adversaries that attempt to discover the hidden states of other users while being honest to the FL protocol. Only the received information such as the global model's parameters is visible to the attackers.
\item \textbf{Training Manipulation} is the process of learning, affecting, or distorting the FL model itself \cite{pmlr-v20-biggio11}. The attacker can damage the integrity of the learning process by attacking the training data or the model during the training phase \cite{pmlr-v97-bhagoji19a}.
\item \textbf{Inference Manipulation} mainly consists of evasion or inference attacks \cite{10.1145/1128817.1128824}. They usually deceive the model into making incorrect decisions or gather information regarding the model's properties. These opponents' efficacy is determined by the amount of knowledge given to the attacker, which classifies them into white-box and black-box variations.
\end{itemize}

\subsection{Attacks on Performance}
\begin{figure*}[h]
\centering
\includegraphics[trim={0.6cm 0.2cm 0.6cm 0.2cm},clip,width=\textwidth]{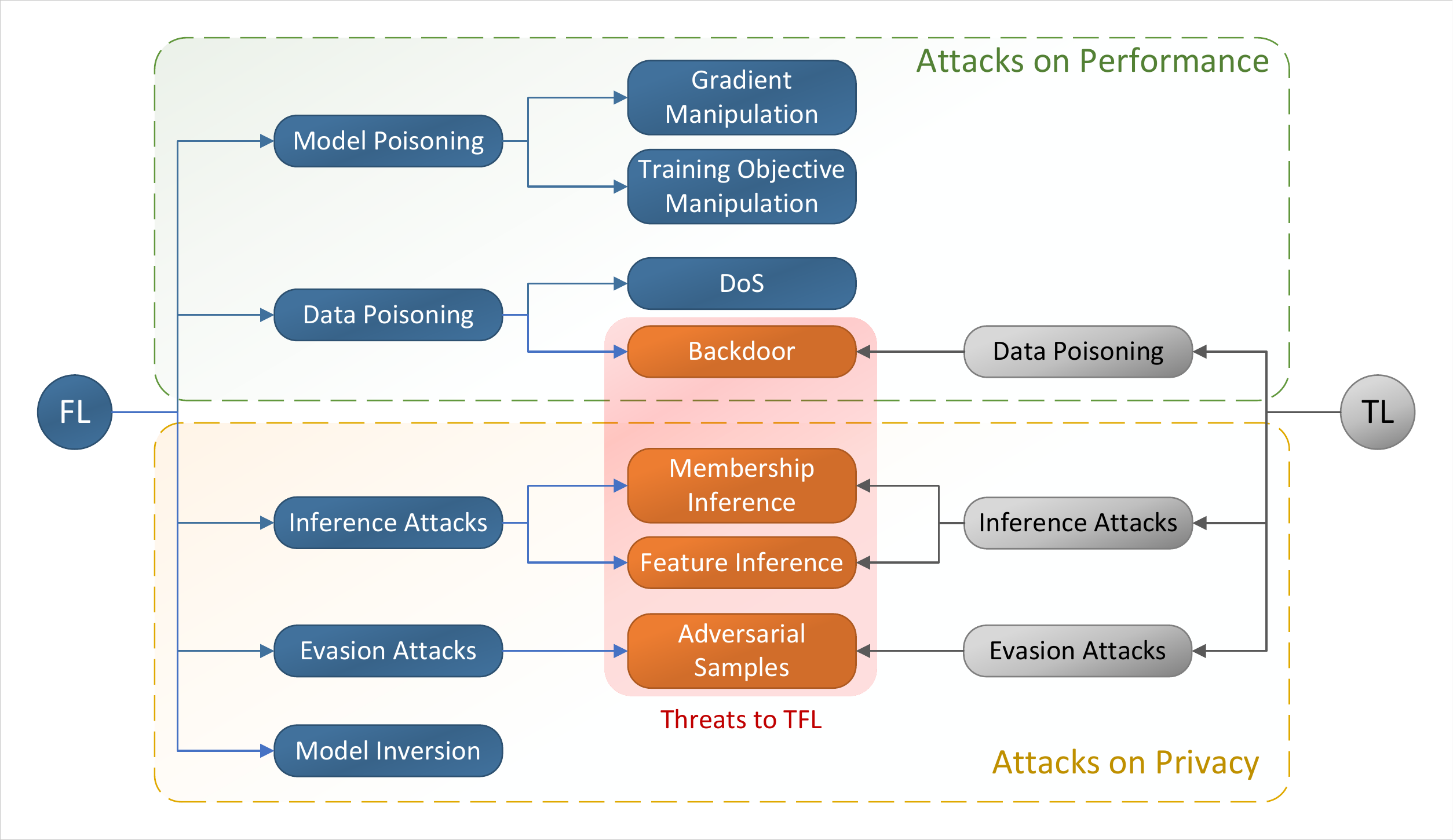}
\caption{Taxonomy of the attacks on Federated and Transfer Learning. The common threats between FL and TL are specified in red area.}
\label{fig:attacks}
\end{figure*}

Figure \ref{fig:attacks} shows the taxonomy of attacks on FL and TL. The common threats between FL and TL that can jeopardize FTL are also specified within the red area.

\subsubsection{Data Poisoning}
Poisoning training data in FL often affect the integrity of the training data, compromising model performance by injecting a backdoor for particular triggers during inference and degrading the overall model accuracy. The most common types of data poisoning attacks are as follows:
\begin{itemize}
\item \textbf{Denial-of-Service (DoS)} attacks generally attempt to minimize the target overall performance, impacting the recognition rate of all classes. In a FL system, label noise in the training data may be induced to create a poisoned model that cannot accurately predict any of the classes. This is also referred to as \textit{label flipping} \cite{DBLP:journals/corr/abs-1912-04977} in the literature. When the parameters of this model are sent to the server, the performance of other devices diminishes.
\item \textbf{Backdoor} attacks are designed to impose intentional and particular false predictions for specific data patterns. \cite{DBLP:journals/corr/abs-1912-04977}. Backdoor attacks, unlike DoS attacks, only influence the model's recognition ability for a specific group of samples or classes.
\end{itemize}

\subsubsection{Model Poisoning}
Poisoning a model refers to a wide range of techniques for tampering with the FL training procedure. It is worth mentioning that in some literature data poisoning is categorized as a type of model poisoning \cite{9308910}. However, here, we mainly target gradient and learning objective manipulation when referring to model poisoning.
\begin{itemize}
\item \textbf{Gradient Manipulation:} Local model gradients may be manipulated by adversaries to degrade overall performance of the central model, for example, by lowering detection accuracy \cite{DBLP:journals/corr/abs-1912-04977}. For instance, this approach is used to inject hidden global model backdoors \cite{pmlr-v108-bagdasaryan20a}.
\item \textbf{Training Objective Manipulation:} involve manipulating model training rules \cite{DBLP:journals/corr/abs-1912-04977}. Training rule manipulation, for example, is used to successfully carry out a covert poisoning operation by appending a deviating term to the loss function to penalize the difference of benign and malicious updates \cite{pmlr-v97-bhagoji19a}.
\end{itemize}

\subsection{Attacks on Privacy}
\subsubsection{Model Inversion Attacks}
It has been demonstrated that model inversion attacks can successfully define sensitive characteristics of the classes and instances covered by the model \cite{DBLP:journals/corr/abs-1912-04977, 9308910}. \cite{10.1145/2810103.2813677} states that using these attacks in a white-box setting on decision trees enables reveal sensitive variables such as survey responses, which may be identified with no false positives. Another study demonstrates that a hacker may anticipate genetic data of a person simply using their demographic information \cite{DBLP:journals/corr/abs-1912-04977}.

\subsubsection{Membership Inference Attacks}
Membership inference aims at disclosing the membership of a particular sample to the training dataset or a certain class. Furthermore, this form of attack can work even if the objective is unrelated to the basic characteristics of the class \cite{8835269}.

\subsubsection{GAN Reconstruction Attacks}
Model inversion is similar to GAN reconstruction; however, the latter is substantially more potent and have been demonstrated to create artificial samples that statistically resemble of the training data \cite{10.1145/3133956.3134012}. Traditional model inversion approaches utterly fail when attacking more sophisticated DL architectures, whereas GAN reconstruction attacks may effectively create desirable outputs. It has been shown that even with the presence of differential privacy, GAN may be able to reach the objective. As a result, an adversary may be able to persuade benign clients to mistakenly commit gradient modifications that leak more confidential details than planned during collaborative learning \cite{10.1145/3133956.3134012}.

\section{Threats to Transfer Learning}

\subsection{Backdoor Attacks}

Plenty of the pre-trained Teacher models ($\mathcal{T}$) employed for TL are openly available, making them vulnerable to backdoor attacks \cite{9112322, 10.1145/3319535.3354209}. In a white-box setup, the intruder has access to $\mathcal{T}$, as is prevalent in modern applications. The intruder intends to cause a erroneous decision making for a Student model ($\mathcal{S}$) that has been calibrated via TL using a publicly available pre-trained $\mathcal{T}$ model.

The attacker may breach the publicly accessible pre-trained $\mathcal{T}$ model prior to the $\mathcal{S}$ system deployment phase. Because the regulations of third-party platforms that store diverse $\mathcal{T}$ models are often inadequate, the platforms contain multiple variations of the same pre-trained neural networks. Since weights of a neural network are not self-explanatory, distinguishing damaging models from refined models is complicated, if not impossible. In this situation, we suppose that the intruder is familiar with the structure and parameters of the $\mathcal{T}$ and has black-box access to $\mathcal{S}$, but is unaware of the specific $\mathcal{T}$ who trained this model and which layers were fixed for training $\mathcal{S}$ \cite{10.5555/3277203.3277300}.

Adversaries might potentially get around the fine-tuning technique by leveraging openly accessible pre-trained $\mathcal{T}$ models to construct $\mathcal{S}$ models. The $\mathcal{S}$ models must be optimized using particular $\mathcal{T}$ models, in which a portion of the $\mathcal{T}$ structure must be incorporated and retrained frequently. In a white-box setting, we presume the intruder knows the certain $\mathcal{T}$ that trained $\mathcal{S}$ and which layers were unchanged throughout the $\mathcal{S}$ training. The adversary, in particular, has access to the architecture and weights of the $\mathcal{S}$ model and may change them.

\subsection{Adversarial Attacks}
In contrast to conventional adversarial attacks, which optimize false data to be mistaken for benign samples, the central notion of adversarial attacks against TL is to optimize a data matrix to imitate the intrinsic representation of the target data. Models transferred by re-learning the last linear layer have recently been shown to be sensitive to adversarial instances produced exclusively using a pre-trained model \cite{rezaei2020targetagnostic}. It has been demonstrated that such an attack can fool models that have been transported with end-to-end fine-tuning \cite{DBLP:journals/corr/abs-2002-02998}. This discovery raises questions about the security of the extensively employed fine-tuning approach.

\subsection{Inference Attacks}
An inference attack resorts to data analysis to gather unauthorized information about a subject or database. If an attacker can confidently estimate the true worth of a subject's confidential information, it can be termed as leaked. The most frequent variants of this approach are membership inference and attribute inference.

\subsubsection{Membership Inference}
The goal of membership inference in machine learning is to establish whether a sample was employed to train the target model. Discovering the membership status of a particular user data might lead to serious information theft \cite{zou2020privacy}. For instance, revealing that a patient's medical records were utilized to train a model linked with an illness might disclose that the patient has the condition.

In contrast with conventional machine learning, there are two attack surfaces for membership inference in TL setting, that is discovering the membership status of samples for both $\mathcal{S}$ and $\mathcal{T}$ models. Furthermore, depending on the abilities of certain adversaries, access to either the $\mathcal{T}$ or $\mathcal{S}$ model may be possible. Given both attack surfaces and the extent of attackers' access to the models, there are three possible attack scenarios:

\begin{enumerate}
    \item The attackers can observe $\mathcal{T}$ and aim at ascertaining the state of the $\mathcal{T}$ dataset's membership. This approach is analogous to the traditional membership inference attack, in which the target model is trained from the ground up.
    \item The $\mathcal{S}$ model is visible to the attackers, and they attempt to ascertain the status of the $\mathcal{T}$ dataset's membership The target model is not directly trained from the target dataset in this scenario.
    \item The attackers observes the $\mathcal{S}$ model and try to deduce the $\mathcal{S}$ dataset's membership status. In contrast to the first scenario, here the target model is transferred from the $\mathcal{T}$ model.
\end{enumerate}

\subsubsection{Feature Inference}
An adversary with partial prior knowledge of a target's record can devise feature inference to fill in the missing features by monitoring the model's behaviour \cite{yeom2018privacy, 10.5555/2671225.2671227, 10.1145/2810103.2813677}. For instance, a description of attribute inference attack is given in \cite{yeom2018privacy} and it has been demonstrated that by incorporating membership inference as a subroutine, this attack may deduce missing attribute values. Based on the missing attributes, a set of distinct feature vectors are generated and passed to the membership inference adversary as input. The output of this process is attribute values that correspond to the vector whose membership is confirmed via membership inference. Experimental validations for the effectiveness of an attribute inference of regression models are also available \cite{yeom2018privacy}.

\section{Defense Mechanisms}
\begin{figure*}[t]
\centering
\includegraphics[trim={0.6cm 0.2cm 0.6cm 0.2cm},clip,width=0.9\textwidth , page=2]{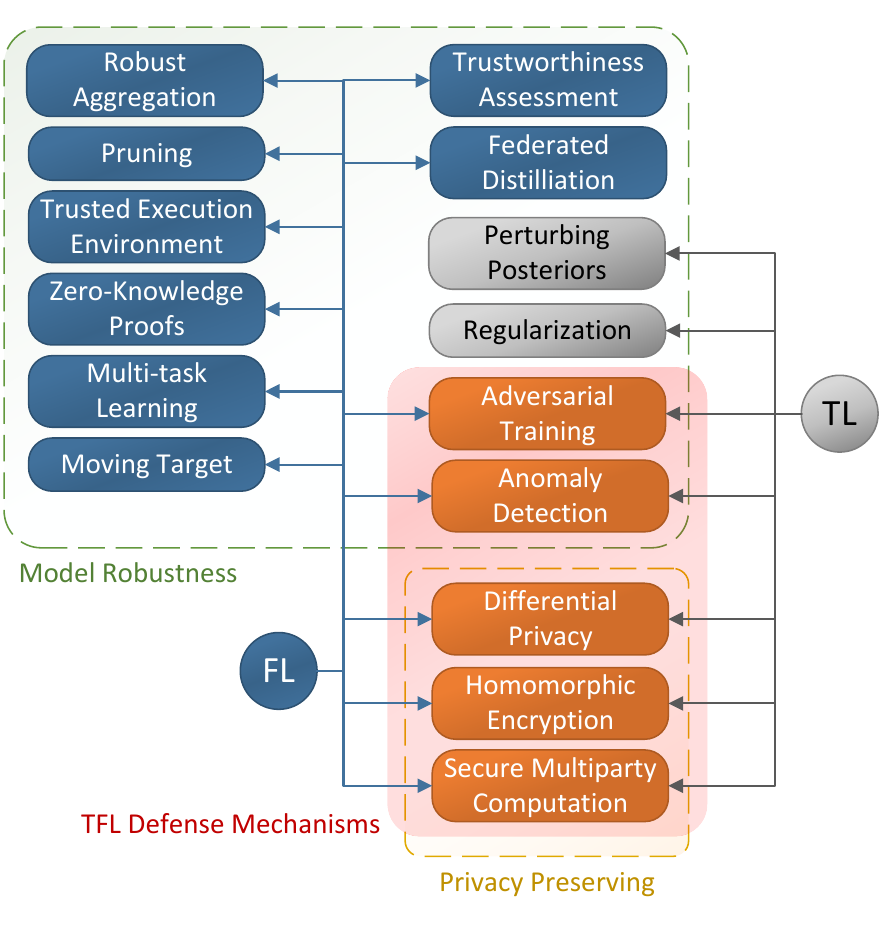}
\caption{Taxonomy of the defense mechanisms for FTL. The common defense mechanisms between FL and TL are specified in red area.}
\label{fig:def}
\end{figure*}

Various defense mechanisms are proposed to fortify FL against privacy and performance related threats. Figure \ref{fig:def} illustrates the taxonomy of the defense mechanisms in TL and FL, and the common approaches between the two that can be used to defend FTL.

\subsection{Privacy Preserving}
Despite the wide diversity of previous efforts on safeguarding FL privacy, suggested methods typically fall into one of these categories: homomorphic encryption, secure multiparty computation, and differential privacy. The following paragraphs go through each of these groups.

\subsubsection{Homomorphic Encryption}
By processing on cyphertext, homomorphic encryption is commonly used to secure the learning process. Clients can use homomorphic encryption to perform arithmetic operations on encrypted data (i.e., cyphertext) without having to decode it. The most prevalent techniques in Homomorphic encryption are explained as in the following \cite{OGBURN2013502}. 

Fully homomorphic encryption is capable of doing arbitrary calculations on the encrypted data \cite{10.1145/1536414.1536440}. This is while partially homomorphic encryption can only execute one operation (e.g., addition or multiplication), and substantially homomorphic encryption can do several operations \cite{10.1007/978-3-642-32009-5_38, 10.1145/359340.359342, 10.1007/3-540-48910-X_16}. The latter, on the other hand, has a restricted amount of additions and multiplications. While completely homomorphic encryption offers greater flexibility, it is inefficient when compared to other forms of homomorphic encryption \cite{10.1145/1536414.1536440}. 

Despite the benefits of homomorphic encryption, executing arithmetic on the encrypted integers increases the memory and processing time costs. For this reason, one of the main problems in homomorphic encryption is to find a proper balance between privacy and utility \cite{10.1145/2857705.2857731, info:doi/10.2196/medinform.8805}. In \cite{8241854}, for instance, additive homomorphic encryption is used to secure distributed learning by securing model changes and maintaining gradient privacy. Another example is \cite{DBLP:journals/corr/abs-1711-10677}, which uses an additive homomorphic architecture to defeat honest-but-curious adversaries using federated logistic regression on the encrypted vertical FL data. However, the overburdening of the system with additional computational and communication costs is a typical downside of such systems.

\subsubsection{Secure Multiparty Computation}
Secure Multiparty Computation (SMC) \cite{4568388} is a sub-field of cryptography, in which multiple parties cooperate to estimate a function on their input, without compromising privacy between participants. As an example of SMC is proposed in \cite{7958569}, which enables collaborative training without compromising privacy. Nevertheless, SMC is followed by considerable computational and communication burden, which may deter parties from collaborating. This dramatic rise in communication and processing costs makes SMC undesirable for large-scale FL.

For safe aggregation of individual model updates, \cite{10.1145/3133956.3133982} suggested a protocol based on SMC that is secure, communication-efficient, and failure-resistant. Their technique makes the communicating information perceivable only when they are aggregated. Thus, their protocol is secured in honest-but-curious and malicious setups. In other words, none of the participants learns anything beyond the aggregate of the inputs of numerous honest users \cite{10.1145/3133956.3133982}.

Aside from the efficiency-related issues of SMC, another major problem for SMC-based systems is the necessity for all participants to coordinate at the same time during the training process. In practise, such multiparty interactions may not be ideal, especially in FL contexts where the client-server design is typical. Moreover, while the privacy of client data is preserved, malicious parties may still infer sensitive information from the final output
\cite{7286780, 10.1145/3196494.3196522}. As a result, SMC cannot guarantee information leakage protection, necessitating the incorporation of additional differential privacy mechanisms within the multiparty protocol to overcome these issues \cite{10.1145/1807167.1807247,10.1007/978-3-642-24178-9_9}. In addition, all cryptography-based protocols preclude the audition of updates, during which a hacker can covertly inject backdoor features into the shared model \cite{pmlr-v97-bhagoji19a}.

\subsubsection{Differential Privacy}
The idea of Diffrential Privacy (DP) is to inject random noise into the generating updates so that the data interpretation becomes infeasible for malicious entities. DP is primarily used to safeguard DFL communications against privacy attacks (e.g., inference attacks); however, the literature also shows that DP is also beneficial against data poisoning, as these attacks are usually designed based on the communicated gradients \cite{10.5555/3367471.3367701, 10.1145/2976749.2978318, DBLP:journals/corr/abs-1712-07557}.

In contrast to homomorphic encryption and SMC whose main disadvantage was communication overhead, DP does not overburden the system in this sense. Instead, DP comes at the cost of deteriorating the model quality since the injected noise can potentially add up to the noise within the constructed model. Moreover, DP provides resistance to poisoning attempts due to its group privacy trait. As a result, as the number of attackers increases, this defense will reduce significantly.

DP can be centralized, local, or distributed. In centralized DP, the noise addition is performed via a server, which makes it impractical in FDL. On the other hand, local \cite{6736718} and distributed DP \cite{10.1145/2873069, dwork2006our} both assume that the aggregator is not trusted, which perfectly complies with the FDL paradigm. In the local variant, participants inject noise to their estimated gradients before sharing them over the blockchain. However, research on local DP indicates its impotency to provide privacy guarantee on large-scale and heterogeneous models with numerous parameters \cite{DBLP:journals/corr/abs-2009-05537,papernot2017semisupervised}. In FDL, the injected noise should be calibrated to ensure successful DP. Despite the appealing security qualities of local DP, its practicality becomes questionable when dealing with immense number of users.

It is also possible to integrate TL into DP. For instance, private aggregation of $\mathcal{T}$ ensembles \cite{papernot2017semisupervised, papernot2018scalable} initially training an ensemble of $\mathcal{T}$s on disjoint subsets of private data, then perturbs the ensemble's information by introducing noise to the aggregated $\mathcal{T}$ votes before transferring the information to a $\mathcal{S}$. The aggregated output of the ensemble then is used to train a $\mathcal{S}$ model, which learns to precisely replicate the ensemble. To meet the desired accuracy, this method requires a large number of clients, and each of them must have sufficient training records. On the other hand, most industrial applications deal with imbalanced data \cite{8892480}, and similarly, FL data is often imbalanced among parties that does not comply with this assumption.

It has been established that the usage of DP helps prevent inference attacks in TL \cite{10.1145/2976749.2978318,236254}, albeit at the cost of potential utility loss \cite{236254,10.1145/2976749.2978355}. By definition, DP seeks to conceal the presence or absence of a record in a dataset, which works against the objective of membership inference attacks. \cite{10.1145/2508859.2516686} draws attention to the fact that these two concepts seem to counteract each other and establishes a link between DP and membership inference attacks. This has been often carried out by minimizing the bias of the model towards any individual sample or feature by including adequate differential privacy noise. The existing connection between records and features is elaborated in \cite{yeom2018privacy}.

\subsection{Model Robustness}
Defenses are classified into two types: proactive and reactive. The former is an inexpensive method of anticipating attacks and associated consequences. The reactive defense operates by detecting an invasion and taking preventative steps. In the production environment, it is often deployed as a patch-up. FL presents multiple additional attack surfaces throughout training, resulting in complicated and unique countermeasures. In this part, we will look at some of the most common types of FL defensive tactics and investigate their usefulness and limits.

\subsubsection{Anomaly Detection}

Anomaly detection methods actively identify and stops malicious updates from affecting the system \cite{Hallaji2022, article}. These methods may be also used in FL systems to identify potential threats \cite{9411833}. One frequent technique for handling untargeted adversaries is to calculate a specific test error rate on updates and reject those disadvantageous or neutral to the global model \cite{article}.

In \cite{10.1145/2991079.2991125}, a protection mechanism is proposed that clusters participants based on their submitted suggestive attributes to identify malicious updates. It produces groups of benign and malicious users with each indicator attribute. Another detector monitors drifts in updates using a distance measure for different participants \cite{NIPS2017_f4b9ec30}. \cite{DBLP:journals/corr/abs-1910-09933} proposed producing low-dimensional model weight surrogates to recognise anomalous updates from participants. An outlier detection-based paradigm presented in  \cite{8418594} selects a number of updates that work in favour of objective function among others. DL-based anomaly detection is often performed using autoencoders \cite{app8122663}. These neural network models represent data in a latent space, in which anomalies can be discriminated. Examples of anomaly detection in FL are given in \cite{247652, li2020learning}.

Backdoor attacks in TL may also be mitigated via anomaly detection. As an example, \cite{8119189} employs an anomaly detection technique to determine whether the input is a possible Trojan trigger. If the input is identified as an anomaly, it will not be passed to the neural network. This approach employs support vector machines and decision trees to find anomalies.

\subsubsection{Robust Aggregation}
The security of FL aggregation techniques is of paramount importance. Extensive research endeavours has been dedicated to research on robust aggregation that can recognize and dismiss inaccurate or malicious updates during training \cite{pillutla2019robust, DBLP:journals/corr/abs-2009-08294}. Furthermore, strong aggregation approaches must be able to withstand communications disturbances, client dropout, and incorrect model updates on top of hostile participants \cite{9026922}. Existing constraints \cite{DBLP:journals/corr/abs-1912-04977} of aggregation methods for integration with FL lead to the emergence of more mature techniques such as adaptive aggregation, have been developed. This technique incorporates repeated median regression into an iteratively re-weighted least squares \cite{DBLP:journals/corr/abs-1912-11464} and a resilient aggregation oracle \cite{pillutla2019robust}. This form of aggregation has been shown to be resistant to distortion rates distortion up to fifty percent of the users. To assess participants' prospective contributions, \cite{10.1145/3432291.3432303} recommends employing a Gaussian distribution. They also provided layer-by-layer optimization procedures to ensure that the aggregation works effectively. Experiments reveal that this aggregation method surpasses the well-known FedAvg in terms of robustness and convergence. Aggregation methods can also help with the problem of FL client heterogeneity. FedProx was designed as a re-parametrization and generalisation of FedAvg \cite{DBLP:journals/corr/abs-1812-06127}. In comparison to FedAvg, it exhibits substantially more consistent and accurate convergence behaviour in highly heterogeneous FL systems.

Pruning also eliminates backdoors in TL by removing duplicate neurons that are no longer relevant for normal classification \cite{10.1007/978-3-030-00470-5_13}. However, it has been discovered that when applied to particular models, it significantly degrades the model performance \cite{8835365}.

\subsubsection{Pruning}
Pruning decreases the size of a deep learning model by removing neurons in order to reduce complexity, increase accuracy, and eliminate backdoors Clients in the FL environment are abundant, and they are frequently linked to the server via unreliable or costly connections. When it comes to training large-scale deep neural networks (DNN), engineers encounter a huge challenge due to the restricted processing power on some edge devices. Federated dropout \cite{DBLP:journals/corr/abs-1812-07210} demonstrates that a good generalization can be achieved by allowing users to perform partial training on the global model. Both transmission and local processing costs are reduced by means of federated dropout. Discarding inactive nodes of a network also make it robust against backdoors \cite{10.1007/978-3-030-00470-5_13}. Passing benign and malicious behaviour into the same set of activations, one can combine pruning with fine-tuning. It has been shown that using this approach the backdoor task accuracy is reduced to zero in several circumstances.

\subsubsection{Trusted Execution Environment}
Trusted Execution Environment (TEE) secures linked devices in FL, establishing digital trust \cite{9411833}. By using an isolated and encrypted part of the main processor, it safeguards devices from inserting incorrect training results. TEE can be used in FL to mitigate algorithmic threats \cite{CHEN202069, 10.1145/3458864.3466628}. The validity of a participating device in a TEE Authentication should be checked by the connected service with which it is attempting to enroll. Furthermore, until the matching party provides a message, the status of code execution stays hidden. The execution route of the code cannot be changed until it takes explicit input or a validated interruption. The TEE is in charge of all data access privileges. Cryptographic technologies are used to secure TEE communications. Only the TEE secure environment stores, maintains, and uses private and public encryption keys. The TEE can show a remote client what code is presently being executed as well as the starting state. TEE can aid in resolving a key challenge for FL security since it is becoming progressively important in securing the central server and clients against hackers and preventing data theft.

\subsubsection{Zero-Knowledge Proofs}
Zero-knowledge proofs allow one party to verify assertions made by another party without exchanging or exposing underlying data \cite{6547113}. In the mid-1980s, MIT researchers initially promoted the notion of zero-knowledge proofs \cite{doi:10.1137/0218012}. Zero-knowledge procedures are probabilistic evaluations, meaning they cannot guarantee something with 100 percent certainty that it will be discovered. Instead, they supply unconnected bits of information that might add up to suggest that an statement's truth is overwhelmingly likely. Thus, zero-knowledge proofs offer a practical answer to the problem of private data verifiability. For instance, zero-knowledge proofs can be employed in FL to make sure the clients' model used authentic feature for training and generating an update. Even though this approach has many appealing potentials for transforming secure update monitoring, we need to better understand how to use these approaches and discover problems in how the modules are constructed and deployed. Zero-knowledge proofs protocols mostly maintain their performance regardless of the volume of data.

\subsubsection{Adversarial Training}
Adversarial training denotes a min-max optimization problem in which the adversarial samples and model parameters are updated alternately. Generally, adversarial samples are generated through maximizing a classification loss, and model parameters are attained via minimizing a loss w.r.t. the generated adversarial samples \cite{DBLP:journals/corr/abs-1812-06127, book, 9609642}. This approach can provide an acceptable resilience against evasion attacks \cite{NEURIPS2019_7503cfac, DBLP:journals/corr/abs-1812-03411}. While there are different approaches to carry out adversarial training, including the so-called generative adversarial networks \cite{Farajzadeh-Zanjani2022,9563211, 9360878, FARAJZADEHZANJANI2021101}, non of them are flawless. To begin with, this approach was mainly designed for independent and identically distributed data. This is while FL data do not comply with this assumption, and, thus, further research is required to investigate the practicality of adversarial training in FL \cite{DBLP:journals/corr/abs-1903-10484}. Furthermore, this approach can be very time-consuming. In addition, adversarial training often improves resilience for cases utilized during the training. Furthermore, it can possibly exhaust FL participants' limited computational capabilities and leaving the trained model exposed to various forms of adversarial noise \cite{pmlr-v97-engstrom19a, NEURIPS2019_5d4ae76f}.

Adversarial training also aids in the prevention of TL inference attacks \cite{zou2020privacy, 10.1145/3243734.3243855}. For example, \cite{10.1145/3243734.3243855} presents a technique for training models with membership privacy, which assures that a model's predictions are indistinguishable on both training and unobserved samples of similar distributions. This technique formulates a min-max problem and develops an adversarial training procedure that minimizes the model's prediction loss along with the attack maximum gain. This method, which ensures membership privacy, also functions as a powerful regularizer and aids in model generalization.

Mitigating adversarial attacks is another use-case of adversarial training. As an example, \cite{aithal2021mitigating} investigates the approach of introducing white noise to DL results to counter these assaults and emphasises on the noise-cost balance. The query count of the attacker is calculated analytically based on the noise standard deviation. Consequently, the degree of noise required to prevent attacks can be easily determined while maintaining the appropriate extent of security defined by query count and limiting performance deterioration.

\subsubsection{Multi-Task Learning}
The statistical and system difficulties of FL such as efficiency and fault fault tolerance are addressed using Federated Multi-task Learning (FML) \cite{NIPS2017_6211080f, pmlr-v139-li21h}. The goal of FML is to learn models for numerous related activities at the same time. It can perfectly handle statistical problems since it can immediately infer associations among non-i.i.d. and imbalanced data. For instance, \cite{NIPS2017_6211080f} designs a FML approach to accelerate convergence while managing devices that disconnect on a regular basis. This approach is also flexible against data heterogeneity.

\subsubsection{Moving Target Defense}
Moving target defense \cite{6578785, 10.1145/2663474.2663486, article2, 9047923} confuses malevolent adversaries by constantly re-configuring the system and make it harder for intruders to infer system states. This may be accomplished by randomly shifting the FL system's components and nullify their knowledge of the system. This defense mechanism also creates complexity and expense for attackers and reduces the disclosure of vulnerabilities and the possibility of an attack. It also improves the system resilience, specially against sniffing attacks. This dynamic mechanisms disables intruders to make accurate estimations regarding the required resources for attacking the FL training process.

\subsubsection{Client Trustworthiness Assessment}
Poisoning attacks in FL are mostly studied in a centralized context. Only a limited number of research endeavours, however, address these attacks in decentralized systems \cite{Xie2020DBA:}, where several adversarial parties follow the same objective and attempt to poison the training data. Although these attacks pose a greater risk in FL, their efficiency remains unknown compared to their centralized variants. This protection approach works by detecting authorized clients and drastically increasing the rate of failure for poisoning attacks, even when the attack is initiated in a distributed fashion.

\subsubsection{Federated Distillation}
Exchanging model parameters becomes prohibitively expensive when communication resources are limited, especially for contemporary big DNNs. In this sense, federated distillation \cite{DBLP:journals/corr/abs-2009-05537} is an appealing FL option since it only transmits model outputs, which are often considerably less in size than the model sizes. Knowledge distillation is a fundamental algorithm in federated distillation \cite{li2019fedmd}. The goal of knowledge distillation is to perform TL from a large model ($\mathcal{T}$) to a compact model ($\mathcal{S}$). In FL, this idea translates into sharing the knowledge of a model rather than the parameters, which improve FL's resilience while reducing communication and computing costs.

\subsubsection{Regularization}
Classifiers make more confident predictions when confronted with data samples they have been trained on before. For this reasons, overfitting of a model can lead to a successful membership inference. Classifiers make more reliable predictions when confronted with records they have been trained on before. To tackle this issue, researchers have investigated the usage of regularization for preventing overfitting, which in turn eliminates membership inference. The conventional $L_2$ regularizer, as an example, is examined during the training process of a target classifier \cite{7958568}.

Dropout is another regularization strategy intended to counter membership inference attacks. It was employed in \cite{DBLP:journals/corr/abs-1806-01246} to counter membership inference attacks. In each training cycle, dropout dismisses a neuron with a particular probability.

Model stacking is a conventional ensemble approach that combines the findings of multiple weak classifiers to form a strong model. \cite{DBLP:journals/corr/abs-1806-01246} investigated the use of this method to counter membership inference. The target classifier, in particular, is comprised of three models grouped into a tree structure (i.e., one model at top and two models at the bottom of the tree). 

The original data samples are fed to leafs of the tree, while the results obtained from the leafs are inputs to the top of the tree. The tree models are trained using separate sets of data samples, which decreases the likelihood that the target classifier will recall any particular point, which in turn reduces overfitting.

\subsubsection{Perturbing Posteriors}
Rather than meddling with the target classifier's training procedure, one can introduce noise to the classifiers' outputs \cite{10.1145/3319535.3363201}. This concept is referred to as perturbing posteriors. For instance, \cite{10.1145/3319535.3363201}, designs a method to protect against membership inference launched in a black-box setting. This defensive approach operates in two stages and offers theoretical robustness guarantee. The first stage involves locating a generated noise vector that may be used to convert a vector of confidence scores into an adversarial example. This noise vector is added to the confidence scores with a probability in the next stage.

\section{Future Research}
As mentioned previously, it is anticipated that FTL is most vulnarible against backdoor, membership inference, feature inference, and adversarial samples (refer back to Fig. \ref{fig:attacks}). In this section, we outline future development requirements that we believe will be promising for FTL in this sense.

\subsection{Decentralized Federated Transfer Learning}
Decentralized FL is a new study topic in which the system has no singular central server. Decentralized FL may be more beneficial in business-based FL instances when third parties are not trusted by the clients. Each client might be selected as a server in a turn-based fashion. As of now, there are no decentralized FTL protocols in the literature. It would be fascinating to see if the same risks that exist in server-based FL also arise in decentralized FTL.

\subsection{Flaws in Current Defense Mechanisms}
Because FL cannot review updates for privacy reasons, it is vulnerable to poisoning attacks, which is often used to counter adversarial attacks in ML, remains a questionable choice in FL since it was designed particularly for i.i.d. data and its effectiveness in non-i.i.d. scenarios is unknown. This can become problematic in the case of FTL furthermore. Besides, adversarial training is computationally expensive and may degrade efficiency, regardless of the type of FL.

Available privacy defenses for FTL are mostly based on homomorphic encryption and secret sharing. Nevertheless, since DP is used as a privacy-preserving method in both FL and TL, it may be also used for FTL. If future works extend DP to FTL, there are a number of points to consider. Firstly, DP cannot handle attribute inference. Secondly, client-level DP is designed for large-scale systems with numerous clients, and using it in smaller systems may affect it performance.

\subsection{Optimizing Defense Mechanism Deployment}
The servers will require additional computational resources while implementing defense mechanisms to verify if any attacker is targeting the FTL system. Additionally, different forms of defense systems may have varying degrees of efficiency against different types of threats, as well as varying costs. It is crucial to look into how to optimize the deployment of defensive systems or the declaration of deterrent measures for FTL.

\subsection{Achieving  Simultaneous Objectives}
There are no extant research on FL or FTL that can achieve the following objectives at the same time \cite{DBLP:journals/corr/abs-2012-06337}:
\begin{enumerate}
    \item Rapid model convergence.
    \item Descent generalization of model.
    \item Efficient communication.
    \item Preserving privacy.
    \item Resilience to targeted and untargeted attacks.
    \item Fault tolerance.
\end{enumerate}
Past efforts sought to tackle several objectives simultaneously \cite{9130840}. \cite{xu2021reputation} tackled cooperative fairness and privacy at the same time, and a architecture has been developed to solve mitigate these problems. To cut communication overhead and provide privacy perks, \cite{10.5555/3327757.3327856} integrated DP with model compression approaches. Another research \cite{bernstein2019signsgd} concentrates on enhancing convergence and preventing gradient leakage. Nevertheless, it is crucial to remember that privacy and robustness are incompatible by nature, as protecting against performance attacks typically necessitates full access to the training samples, which is irreconcilable with FTL's privacy requirements. Even though encryption and DP-based approaches can guarantee verifiably privacy-preserving, they are vulnerable to poisoning techniques and may result in models with unfavourable privacy-performance trade-off. Finding a cohesive design that meets all of the aforementioned criteria is indeed undiscovered in the FTL domain.

\subsection{Heterogeneity of Federated Transfer Learning}
The vast majority of privacy and robustness studies have been conducted on FL with homogenous designs. On the other hand, there is a common assumption of feature co-occurrence among most of the available work on heterogeneous FL. For FTL to be secure, the existing defense mechanisms should be compatible with fully heterogeneous feature space \cite{9005992}.

\section{Conclusion}
FTL is one of the latest fields of machine learning, it is evolving at a rapid pace and will be a focal point of research in machine learning and privacy. As FL and TL evolve, so will the dangers to FTL's privacy and security. It is critical to conduct a wide assessment of present FL and TL threats and countermeasures so that upcoming FTL designs consider the possible weaknesses in existing models. This survey provides a clear and straightforward review of the privacy and robustness attack and possible defense mechanisms that may be used in FTL. Designing a coherent FTL defensive mechanism that can withstand various attacks without decreasing model performance would demand multidisciplinary collaboration in the scientific community.

\section*{Acknowledgements}
This work is supported by the Natural Sciences and Engineering Research Council of Canada (NSERC) under funding reference numbers CGSD3-569341-2022 and RGPIN-2021-02968.

\bibliographystyle{spmpsci}
\bibliography{refs}

\end{document}